\documentclass{article}
\usepackage{ijcai20}

\usepackage{times}
\usepackage{soul}
\usepackage{url}
\usepackage[hidelinks]{hyperref}
\usepackage[utf8]{inputenc}
\usepackage[small]{caption}
\usepackage{graphicx}
\usepackage{amsmath}
\usepackage{amsthm}
\usepackage{booktabs}
\usepackage{algorithm}
\usepackage{algorithmic}
\usepackage{amsmath, amssymb}
\usepackage{multirow}
\title{Fact-based Dialogue Generation with Convergent and Divergent Decoding}

\author{
Ryota Tanaka
\and
Akinobu Lee
\affiliations
Nagoya Institute of Technology\\
\emails
\{rtanaka, ri\}@slp.nitech.ac.jp
}
\begin{document}

\maketitle

\begin{abstract}
Fact-based dialogue generation is a task of generating a human-like response based on both dialogue context and factual texts. Various methods were proposed to focus on generating informative words that contain  facts effectively. However, previous works implicitly assume a topic to be kept on a dialogue and usually converse passively, therefore the systems have a difficulty to generate diverse responses that provide meaningful information proactively. This paper proposes an end-to-end Fact-based dialogue system augmented with the ability of convergent and divergent thinking over both context and facts, which can converse about the current topic or introduce a new topic. Specifically, our model incorporates a novel \textit{convergent and divergent decoding} that can generate informative and diverse responses considering not only given inputs (context and facts) but also inputs-related topics. Both automatic and human evaluation results on DSTC7 dataset show that our model significantly outperforms state-of-the-art baselines, indicating that our model can generate more appropriate, informative, and diverse responses.
\end{abstract}

\section{Introduction}
The popularization of social networking services and the improvement in computer technology offer the potential of open-domain conversational dialogue systems that can be trained in an end-to-end manner without any hand-coding. Specifically, a sequence to sequence (seq2seq) model \cite{DBLP:conf/nips/SutskeverVL14,NCM} is the most common approach in the field of dialogue generation, and it can generate fluent responses. However, dialogue systems which only rely on utterances and responses as training data sometimes generate generic and uninformative responses \cite{DBLP:conf/coling/VougiouklisHS16,DBLP:conf/icann/YeCLHL18,DBLP:conf/acl/HuangKGx18}.

\begin{figure}[htb]
    \centering
    \includegraphics[scale=0.35]{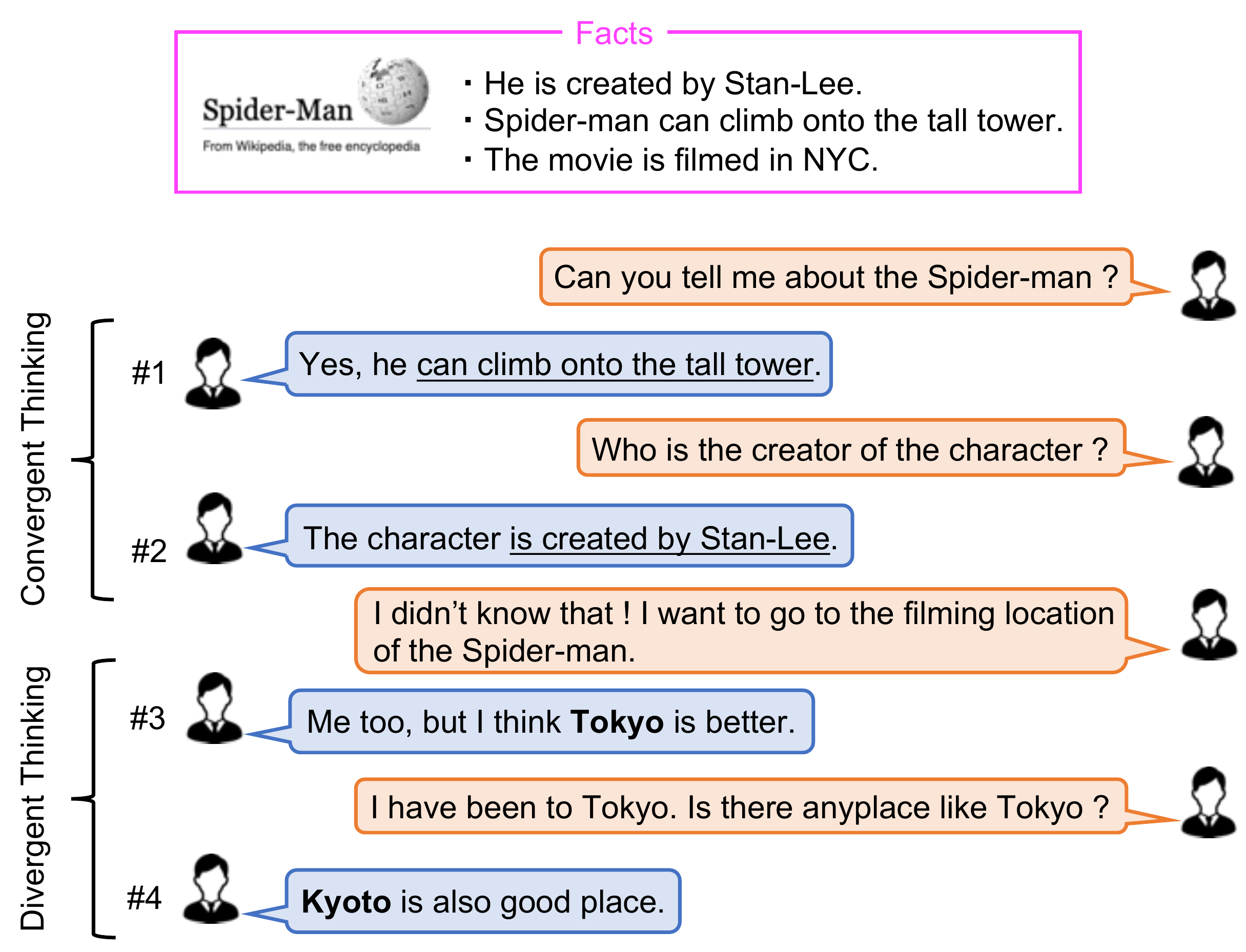}
    \caption{Typical example of fact-based dialogue session. The words with \underline{underline} correspond to the facts. The new topics that are drifted from the candidates of topics are shown in \textbf{bold}.}
    \label{fig:example_intro}
\end{figure}

In order to solve these problems, various works were proposed to generate more human-like responses by exploiting a set of textual facts, such as review posts \cite{DBLP:conf/aaai/GhazvininejadBC18,DBLP:conf/emnlp/MogheABK18}, character description as persona \cite{DBLP:conf/acl/KielaWZDUS18}, and fragments of articles from web-sites \cite{galley2019grounded,DSTC7-SOTA}. These works mainly focus on integrating facts based on a topic of input utterances into dialogue systems. However, this scheme tends to converse passively while keeping the given topic and has a difficulty to bring diverse and proactive communication, as we can often observe in our daily conversation.

To address this issue, this paper makes a step towards creating another type of fact-based dialogue system: endowing it with the ability of a convergent and divergent thinking over both context and facts. This is inspired by human thought method \cite{guilford1967nature}. In general, as shown in Figure \ref{fig:example_intro}, humans usually converses by exploiting available information (i.e., convergent thinking) as well as exploring many possible directions (i.e., divergent thinking) within the interactions. For example, \#1 and \#2 utterances conduct the convergent thinking to talk about the topic of the character \textit{spider-man} by utilizing the facts. Another remarkable phenomenon is that the conversation usually drifts from one topic to other similar topics. Utterances in item \#3 and \#4 perform the divergent thinking to introduce the new topics of \textit{Tokyo} and \textit{Kyoto} by expanding from the topic candidates of \textit{NYC} and \textit{Tokyo} that are present in facts and context, respectively.

This has the potential of enabling the system to chat naturally with humans on open-domain topics and proactively lead the conversation smoothly, which results in more informative and diverse conversation. Motivated by this, we propose a fact-based neural conversational model equipped with a \textit{convergent and divergent decoding}, which can converse about the current topic or drift from one topic to another. Specifically, our model can enable the decoder to attend and copy tokens from source inputs including context and facts (i.e., \textit{convergent decoding}) or inputs-related topics that are produced by a topic drifter (i.e., \textit{divergent decoding}), considering the decision made by a decoding switcher. We test the effectiveness of our proposed model on DSTC7 dataset \cite{galley2019grounded} that is one of the most large-scale and challenging benchmarks based on real-world conversation. Both automatic and human evaluation results show that our method significantly outperforms  state-of-the-art methods in producing more appropriate, informative, and diverse responses. 

The contributions of this paper can be summarized below:
\begin{itemize}
    \item To the best of our knowledge, we first incorporate a concept of convergent and divergent thinking over both context and facts into fact-based dialogue system.
    \item We propose a fact-based dialogue system with a convergent and divergent decoding, which can converse about the current topic or drift from one topic to another. 
    \item Experiments show that our model outperforms state-of-the-art models by a large margin in terms of both automatic and human evaluations.
\end{itemize}

\section{Proposed Method}
\subsection{Task Definition}
The task of fact-based dialogue generation can be stated as follows: given a dialogue context $X^c=\{x^c_1, ..., x^c_I\}$ and a collection of $K$ supporting facts $X^{f_k}=\{x^{f_k}_1, ..., x^{f_k}_J\}(k=1,...,K)$, the goal is to generate a response as a sequence of tokens $Y=\{y_1, ..., y_T\}$. We formalize this setting by following DSTC7 sentence generation task \cite{galley2019grounded}. The length of the context is limited to the past six utterance at most, and these are concatenated by the special symbol $<$SEP$>$ representing a break in the utterance. Moreover, the supporting facts are shown in text fragments extracted from various web-sites based on the topics of conversation. 

\subsection{Model Overview}
\begin{figure}[htb]
    \centering
    \includegraphics[height=6cm, width=8.5cm]{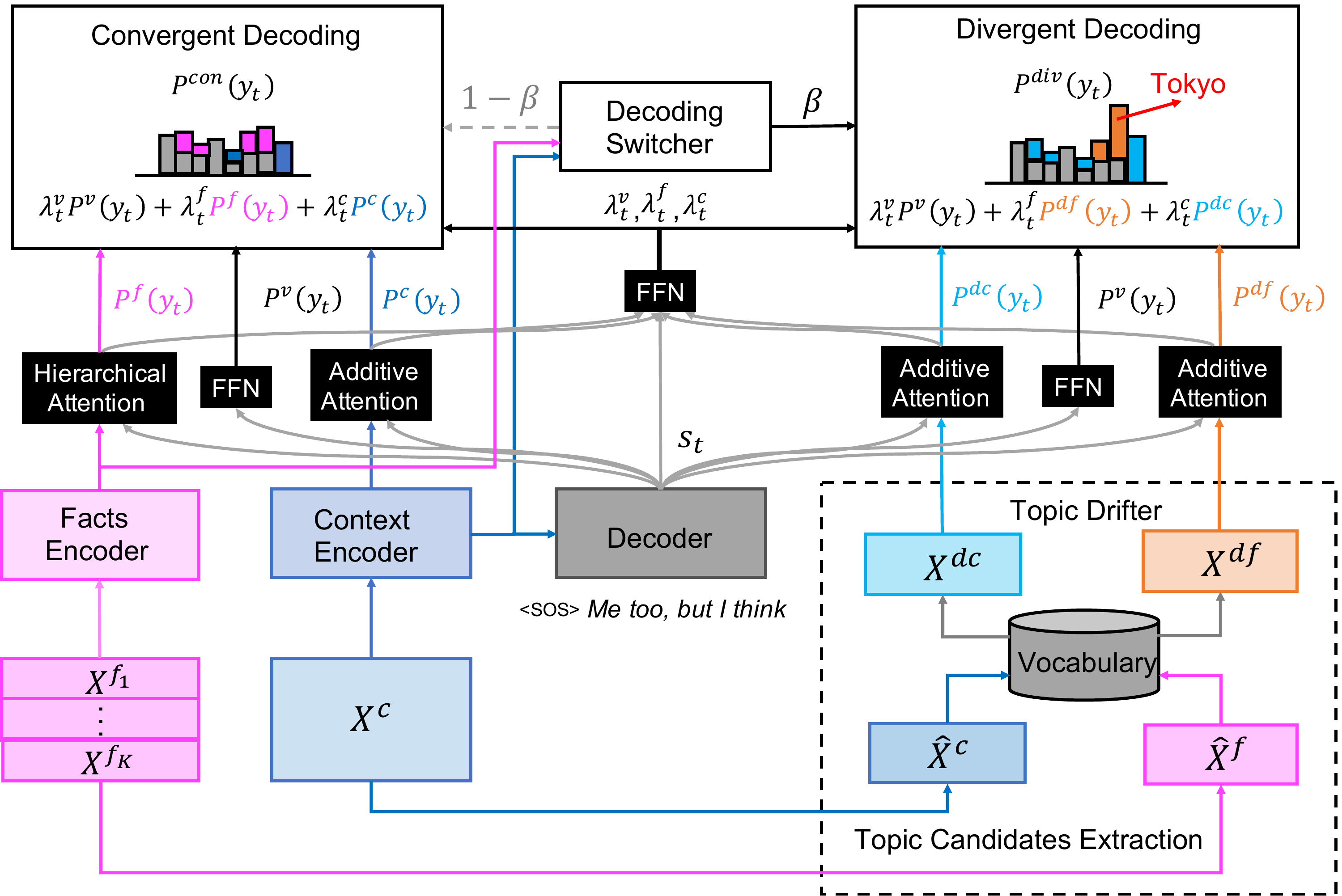}
    \caption{Architecture overview of our model. }
    \label{fig:proposed_model}
\end{figure}
The architecture overview of our proposed model is presented in Figure \ref{fig:proposed_model}, and it consists of four major components:  \textit{context \& facts encoders} (\S \ref{sec: Encoders}), \textit{topic drifter} (\S \ref{sec: Topic Drifter}), \textit{decoding switcher} (\S \ref{sec: Decoding Switcher}), and \textit{decoder} (\S \ref{sec: Decoder}). The context \& facts encoders encode $X^c$ and $X^{f_k}$ into hidden vectors $H^c$ and $H^{f_k}$, respectively. The topic drifter outputs inputs-related topics (drift words) $X^{dc}$ and $X^{df}$ that can be expanded from candidates of current topics in $X^c$ and $X^f$, respectively. In the next utterance, the decoding switcher predicts the probabilities of executing the \textit{convergent decoding} or \textit{divergent decoding}. Based on the decision made by the decoding switcher, the decoder attends and copies tokens from source inputs; context and facts (\textit{convergent decoding}) or drift words that are produced by the topic drifter (\textit{divergent decoding}). 

\subsection{Context and Facts Encoders}\label{sec: Encoders}
We implement the encoder with forward and backward Gated Recurrent Units (GRU) \cite{GRU}. The encoder converts a dialogue context and a collection of facts into two hidden state sequences $H^{c} = \{H^c_1, ..., H^c_I\}$ and $H^{f_k} = \{H^{f_k}_1,..., H^{f_k}_J\}(k=1,2,...,K)$, respectively.
\begin{eqnarray}
H^c_i = [{\rm GRU}(\overrightarrow{h^c_{i-1}}, e(x^c_i));{\rm GRU}(\overleftarrow{h^c_{i-1}}, e(x^c_i))], \hspace{0.35cm} \\
 H^{f_k}_{j} = [{\rm GRU}(\overrightarrow{h^{f_k}_{j-1}}, e(x^{f_k}_j));{\rm GRU}(\overleftarrow{h^{f_k}_{j-1}}, e(x^{f_k}_j))],
\end{eqnarray}
where the bracket $[\cdot ; \cdot]$ denotes the vector concatenation. Here, $H^c \in \mathbb{R}^{2d \times I}$ and $H^{f_k} \in \mathbb{R}^{2d \times J}$ where $d$ is the hidden size are vectors concatenated with respect to all the forward hidden states $\overrightarrow{h}$ and backward states $\overleftarrow{h}$ as well. Moreover, $e(\cdot)$ is the pre-trained GloVe word embedding \cite{DBLP:conf/emnlp/PenningtonSM14}. To reduce the number of parameters, both encoders share weights and word embeddings.

\subsection{Topic Drifter}\label{sec: Topic Drifter}
Topic words are one of the crucial factors for generating an informative response \cite{DBLP:conf/acl/HuangKGx18}. However, repeating a response based on a certain topic hurts diversity and smoothness of the conversation. To overcome this problem, the topic drifter outputs inputs-related words called drift words to expand from candidates of current topics. 

The topic drifter performs two processes to output the drift words from context and facts. Firstly, candidates of current topic words $\hat{X}^c=\{\hat{x}^c_1,...,\hat{x}^c_{L'}\}$ and $\hat{X}^f=\{\hat{x}^f_1,...,\hat{x}^f_{M'}\}$ are extracted from context and facts, respectively. Specifically, we apply the existing topic words extracter \cite{DBLP:conf/acl/TangZXLXH19} which combines TF-IDF and Part-Of-Speech features for scoring word salience. Secondly, the cosine similarity between a word from vocabulary constructed using the training set and each extracted topic word is applied, and the most related $N_{div}$ words are outputted. After these processes, we finally obtain contextual drift words $X^{dc}=\{x^{dc}_1, ..., x^{dc}_{L}\}$ and factual drift words $X^{df}=\{x^{df}_1, ..., x^{df}_{M}\}$, where $L = L' \times N_{div}$ and $M = M' \times N_{div}$.

\subsection{Decoding Switcher}\label{sec: Decoding Switcher}
Our proposed model has two types of decoding methods for generating a response. The first one is a convergent decoding that can copy tokens from the given inputs including context and facts, and the second one is a divergent decoding that can output a response that contains the drift words obtained by the topic drifter. The decoding switcher selects which the decoding type should be used in the next system utterance. Specifically, the switcher is modeled as a binary classifier that takes a  dialogue context and a set of supporting facts, and it predicts the probability $\beta \in [0,1]$, where $\beta = 0$ means the convergent decoding and $\beta = 1$ means the divergent decoding. The probability of this binary classifier is calculated as:
\begin{eqnarray}
\beta = {\rm sigmoid}(W^\mathsf{T}_{sw} [H^c_{I};H^f_{J}] + b_{sw}),
\end{eqnarray}
where $W_{sw}$ and $b_{sw}$ are learnable parameters and $H^f_{J} = \frac{1}{K}\sum_k H^{f_k}_{J}$, simply the average pooling over all last hidden states of the facts encoder.

Furthermore, we introduce a \textit{switch loss} to better supervise this switcher explicitly. Intuitively, when any facts are not matched the conversation, drifting the topic is necessary (i.e., $\beta$ should be higher) because the incorporation of irrelevant facts may be noisy signals of the model. To this end, we define a correct label $\hat{\beta}$ for each sample. This label is set to 1 if the output word exists in the topic words of the facts and 0 otherwise. The switch loss is defined as follows: 
\begin{eqnarray}
\mathcal{L}_{sw} = - \hat{\beta} \log \beta 
- (1 - \hat{\beta}) \log (1 - \beta). \label{eq:switch}
\end{eqnarray}

\subsection{Decoder}\label{sec: Decoder}
Based on the decision $\beta$ made by the decoding switcher, the decoder generates a response word-by-word sequentially with controlling the output distribution.

The $t$-th hidden state $s_t$ of the decoder is computed autoregressively with the forward GRU where the initial hidden state $s_0$ is $H^{c}_I$ and the initial word $y_0$ is the special symbol $<$SOS$>$ denoting the start of sentence.
\begin{eqnarray}
s_t = {\rm GRU}(e(y_{t-1}), s_{t-1}).
\end{eqnarray}

The probability distribution $P^v(y_t)$ over the fixed vocabulary is obtained by passing the decoder hidden state $s_t$, 
\begin{eqnarray}
P^v(y_t) = {\rm softmax}(W_{v}^\mathsf{T}s_t + b_{v}).
\end{eqnarray}
where $W_{v}$ and $b_v$ are parameters to be learned.

\subsubsection{Multi-Source Attention and Copy Mechanism}
After estimating the $t$-th decoder hidden state $s_t$, we obtain the context vectors $\hat{s}^{c}_{t}$,$\hat{s}^{f}_{t}$,$\hat{s}^{dc}_{t}$,$\hat{s}^{df}_{t}$ and the copy distributions $P^c(y_t),P^f(y_t),P^{dc}(y_t),P^{df}(y_t)$, by leveraging the attention mechanism \cite{DBLP:journals/corr/BahdanauCB14} and the copy mechanism \cite{DBLP:conf/acl/SeeLM17} for the dialogue context, facts, contextual drift words, and factual drift words. 

For the dialogue context, the attention weight $\alpha^{c}_{t}$ and the context vector $\hat{s}^{c}_{t}$ are calculated based on the hidden states $H^c$ obtained from the context encoder and the decoder hidden state $s_t$ (Eq. \ref{eq:(6)}). Then, the copy distribution $P^c(y_t)$ is calculated by integrating the attention weights (Eq. \ref{eq:(7)}).
\begin{eqnarray}
\alpha^{c}_{t}, \hat{s}^{c}_{t} &=& {\rm Attention}(H^c, s_t), \label{eq:(6)} \\
P^c(y_t) &=& \sum_{\{i:x_i=y_t\}}\alpha^{c}_{t,i}. \label{eq:(7)}
\end{eqnarray}

In order not to attend words in irrelevant facts, for calculating the attention of facts, the hierarchical attention mechanism \cite{DBLP:journals/corr/abs-1908-10731} which combines both the word-level and sentence-level attention is applied. First, $\alpha^{f_k}_{t}$ and $\hat{s}^{f_k}_{t}$ are calculated for each word (Eq. \ref{eq:(8)}). Subsequently, $\alpha^{f}_{t}$ and $\hat{s}^{f}_{t}$ are calculated for each fact (Eq. \ref{eq:(9)}). To compute the copy distribution of facts, Eq. \ref{eq:(10)} is used.
\begin{eqnarray}
\alpha^{f_k}_{t}, \hat{s}^{f_k}_{t} &=& {\rm Attention}(H^{f_k}, s_t), \label{eq:(8)} \\
\alpha^{f}_{t}, \hat{s}^{f}_{t} &=& {\rm Attention}([\hat{s}^{f_1}_{t},...,\hat{s}^{f_K}_{t}], s_t), \label{eq:(9)} \\
P^f(y_t) &=& \sum_{k=1}^K \alpha^{f}_{t,k}\sum_{\{j:x^{f_k}_j=y_t\}}\alpha^{f_k}_{t,j}. \label{eq:(10)}
\end{eqnarray}

For the contextual and factual drift words, we compute the attention weights $\alpha^{dc}_{t},\alpha^{df}_{t}$ and the context vectors $\hat{s}^{dc}_{t},\hat{s}^{df}_{t}$ (Eq. \ref{eq:(11)}, \ref{eq:(12)}) based on the embeddings of drift words and the $t$-th decoder hidden state. In addition, the attention weights are summed up to obtain the distribution over the contextual and factual drift words $P^{dc}(y_t)$ and $P^{df}(y_t)$ (Eq. \ref{eq:(13)}, \ref{eq:(14)}).
\begin{eqnarray}
\alpha^{dc}_{t}, \hat{s}^{dc}_{t} &=& \text{Attention}([e(x^{dc}_1), ..., e(x^{dc}_{L})], s_t),\label{eq:(11)} \\ 
\alpha^{df}_{t}, \hat{s}^{df}_{t} &=& \text{Attention}([e(x^{df}_1), ..., e(x^{df}_{M})], s_t), \label{eq:(12)}\\ 
P^{dc}(y_t) &=& \sum_{\{l:x^{dc}_{l}=y_t\}}\alpha^{dc}_{t,l}, \label{eq:(13)}\\
P^{df}(y_t) &=& \sum_{\{m:x^{df}_{m}=y_t\}}\alpha^{df}_{t,m}.\label{eq:(14)}
\end{eqnarray}

\subsubsection{Convergent and Divergent Decoding}
The final output distribution is determined according to the decision $\beta$ made by the decoding switcher. Let $P^{con}(y_t)$ be a output probability of $y_t$  from the convergent decoding, and $P^{div}(y_t)$ be a output probability of $y_t$ from the divergent decoding. Then, these probabilities can be calculated based on the following formula:
\begin{eqnarray}
P^{con}(y_t) &=& \lambda^{v}_t P^v(y_t) + \lambda^{c}_t P^{c}(y_t) + \lambda^{f}_t P^{f}(y_t), \label{eq:con} \hspace{1cm}\\
P^{div}(y_t) &=& \lambda^{v}_t P^v(y_t) + \lambda^{dc}_t P^{dc}(y_t) + \lambda^{df}_t P^{df}(y_t), \label{eq:div} \\
\lambda^v_t, \lambda^c_t, \lambda^f_t &=& 
\text{softmax}(W_{\lambda}^\mathsf{T} [s_t;\hat{s}^{c}_t;\hat{s}^{f}_t;\hat{s}^{dc}_t;\hat{s}^{df}_t]),
\end{eqnarray}
where $W_{\lambda}$ is learnable parameter. 

The final distribution of $y_t$ is defined as the following mixture of the distributions: 
\begin{eqnarray}
P(y_t) &=& \beta P^{div}(y_t) + (1 - \beta)P^{con}(y_t), \\
 &=& \lambda^{v}_t P^v(y_t) + \beta (\lambda^{c}_t P^{dc}(y_t) + \lambda^{f}_t P^{df}(y_t)) \nonumber \hspace{0.5cm}\\
 &+& (1- \beta) (\lambda^c_t P^c(y_t) + \lambda^{f}_t P^f(y_t)), \label{eq:final}
\end{eqnarray}
As can be seen Eq.\ref{eq:final}, the $\beta$ predicted by the decoding switcher only affects the copy distributions in Eq. \ref{eq:con} and Eq. \ref{eq:div} (i.e., not affects the vocabulary distribution). In order to control the output distribution more efficiently, if the output word $y_t $ exists in the copy sources, using the copy distributions preferentially is ideal. To this end, we devise a \textit{copy loss}. Let $\lambda^{cp}_{t} = \lambda^c_t + \lambda^f_t$ be a mixing factor of the $t$-th copy distribution. Then, the copy loss is calculated as follows: 
\begin{eqnarray}
\mathcal{L}_{cp} = - \frac{1}{T} \sum_{t=1}^{T}\{\hat{\lambda}^{cp}_{t} \log(\lambda^{cp}_{t}) + (1 - \hat{\lambda}^{cp}_{t})\log(1-\lambda^{cp}_{t})\},
\end{eqnarray}
where $\hat{\lambda}^{cp}_{t}$ is a correct label defined by checking whether the $t$-th output word $y_t$ exists in the copy sources.

\subsection{Multi-Task Learning}
Our model is trained using minimization of the weighted-sum of three losses $\mathcal{L}_{multi}$ as: 
\begin{eqnarray}
\mathcal{L}_{multi} = \mathcal{L}_{nll} + \gamma_{sw} \mathcal{L}_{sw} + \gamma_{cp} \mathcal{L}_{cp}
\end{eqnarray}
where $\gamma_{sw}$ and $\gamma_{cp}$ are hyper-parameters to balance loss, and $\mathcal{L}_{nll} = - \frac{1}{T} \sum_{t=1}^{T} \log P(y_t)$ is the negative log-likelihood loss in the seq2seq learning \cite{DBLP:conf/nips/SutskeverVL14}.

\section{Experimental Setup}
\subsection{Dataset}\label{sec:dataset}
\begin{table}[ht]
    \begin{center}
    \footnotesize
        \begin{tabular}{c|c|ccc} \hline 
         Subset & & train & dev & test \\ \hline 
         \multirow{2}{*}{Dialogue}
         &\#Pairs  & 1662093 & 110184 & 2208 \\
         &\#Tokens/Utt.  & 20.50 & 21.89 & 19.64 \\ \hline 
         Facts & \#Tokens/Sent. & 45.62 & 45.88 & 45.37 \\ \hline
        \end{tabular}
    \caption{Statistics of DSTC7 dataset.}
     \label{tab:stats}
    \end{center}
\end{table}

We train and evaluate our model on the DSTC7 dataset \cite{galley2019grounded} because it is very large scale and good characterization of real-world conversations. Table \ref{tab:stats} shows the statistical information of the DSTC7 dataset used in our experiment. The dialogue data is crawled from Reddit\footnote{\url{https://www.reddit.com}}, and the facts data is crawled from 226 types of web-sites by using a word matching algorithm. In addition, we follow \cite{DSTC7-SOTA} and use IDF-based knowledge extractor, and a maximum of four sentences are collected as a set of facts. All responses and utterances are truncated to have a maximum length of 32, and maximum fact length are set to 50. 

\subsection{Implementation Details}
The implementation of all the models was conducted with PyTorch\footnote{\url{https://pytorch.org}}. The encoder and decoder have a one-layer GRU structure with 128 hidden states, but these did not share parameters. The word embedding size was set to 300 and the embedding state is initialized using GloVe. During training, the batch size was set to 64 and Adam \cite{adam} with an initial learning rate of 0.0005 was employed. All models were trained at most 10 epochs, and the model with the minimum loss was selected. The vocabulary was obtained from the dialogue and knowledge data, and the size was set to 30,000. The balancing parameters for loss functions $\gamma_{sw}$ and $\gamma_{cp}$ were set to 1.0. The hyper-parameter $N_{div}$ for the topic drifter was set to 5. The label smoothing technique \cite{DBLP:conf/cvpr/SzegedyVISW16} with 0.9 for the binary cross entropy was applied. The accuracy of the decoding switcher achieved 68.7\% on the dev dataset. For inference, top$k$ sampling \cite{DBLP:conf/acl/LewisDF18} with $k=10$ provided the best results for all models\footnote{In a preliminary experiment, de-facto standard decoding methods including beam search, greedy search, and random sampling were also tried for the same conditions.}.
This result is also reported in the previous work \cite{DBLP:conf/acl/QinGBLGDCG19}.

\subsection{Baseline Models}
Several models implemented by ourselves are selected to be evaluated as follows:
\begin{itemize}
    \item \textbf{S2S} is a normal seq2seq that does not have access to the external knowledge \cite{NCM}.
    \item \textbf{kgS2S} is a knowledge-grounded seq2seq model from \cite{DBLP:conf/aaai/GhazvininejadBC18}, where the knowledge is stored in memory units and fed into the decoder.
    \item \textbf{PGN} is a pointer generator network \cite{DBLP:conf/acl/SeeLM17}, where the output distribution is integrated with both the vocabulary and context distributions.
    \item \textbf{kgPGN} is an extended PGN \cite{DSTC7-SOTA} that can also copy from the knowledge. This model won the DSTC7 competition.
    \item \textbf{DeepCopy} is an extended kgPGN model by introducing a hierarchical pointer network. \cite{DBLP:journals/corr/abs-1908-10731}.  
\end{itemize}
In addition to these models, we will also report the automatic evaluation results of the DSTC7 submission models that are taken from the overview paper \cite{galley2019grounded}. 

Furthermore, we also conducted several ablation tests to verify the effectiveness of loss functions $\mathcal{L}_{sw} $ and $\mathcal{L}_{cp}$. Specifically, we set $\gamma_{sw}, \gamma_{cp}$ to 0 during training, respectively. Moreover, we tried to generate different responses by setting the probability $\beta$ in Eq. \ref{eq:switch} to different values in both the training and inference stage. Specifically, we set $\beta$ to 0 (convergent decoding) and 1 (divergent decoding). Then, in order to suppress the impact of $\mathcal{L}_{sw}$ during training, we set $\gamma_{sw}$ to 0.  

\subsection{Evaluation Metrics}
Both automatic and human evaluation metrics are used to analyze the model’s performance. For the automatic evaluation, three types of metrics are used: 1) \textbf{BLEU} is used to measure the 4-gram level \textit{relevance} between the generated responses and the ground-truth. 2) \textbf{Dist-1,2} are used to evaluate the \textit{diversity} of $N$-gram ($N=1,2$) in the generated responses. BLEU and Dist-1,2 are evaluated with the public source code of DSTC7\footnote{https://github.com/mgalley/DSTC7-End-to-End-Conversation-Modeling/tree/master/evaluation}. 3) \textbf{PMI} aims to evaluate how \textit{smoothly} the generated responses are connected to the context and facts. Similar to  \cite{DBLP:journals/corr/abs-1906-09795}, we calculate this metric based on the point-wise mutual information
%\footnote{Let $P_{in}(x)$ and $P_{out}(x)$ be the probabilities that the words will appear in a certain input sentence and response, respectively. Then, $\text{PMI}(x,y)$ is calculated as $\log_2\{ P(x,y)/(P_{in}(x)P_{out}(y))\}$.} 
(PMI) \cite{DBLP:journals/coling/ChurchH90} as follows:
\begin{eqnarray}
\frac{1}{T}\sum_{t=1}^{T} \max_{x \in \{{\rm cont}(X^c), {\rm cont}(X^f)\}} \text{PMI}(x, y_t). 
\end{eqnarray}
where ${\rm cont}(\cdot)$ denotes the content words. 

For the human evaluation, the generated responses are manually assessed based on three metrics: 1) \textbf{Appropriateness} (whether the responses are appropriate/relevant to the given context), 2) \textbf{Informativeness} (whether the responses provide meaningful information relevant to the given context), and 3) \textbf{Fluency} (whether the responses are fluent and could plausibly have been produced by humans). We randomly sampled 100 context-response pairs, shuffled the order of systems, and asked three students who have studied in the field of dialogue systems to rate the pairs in terms of model quality on 0 to 2 scales (2 for the best). In Section \ref{sec:human_evaluation}, the macro-average score will be reported.

\section{Evaluation Results and Analysis}
\subsection{Automatic Evaluation}
\begin{table}[htb]
    \centering
    \fontsize{9pt}{0.4cm}\selectfont
    \begin{tabular}{c|c|cc|c} \hline
        Model & BLEU & Dist-1 & Dist-2 & PMI \\ \hline
        TeamA* & 1.48 & 9.58 & 27.55 & - \\    
        TeamB*$\dag$ & 1.83 & 10.89 & 32.49 & -  \\  
        TeamC* & 1.32 & 5.34 & 17.09 & -  \\  
        TeamD* & 1.35 & 9.37 & 33.37 & -  \\  
        TeamF* & 1.01 & 6.36 & 17.65 & -  \\  
        TeamG* & 1.21 & 3.36 & 26.47 & -  \\  \hline 
        S2S & 1.44 & 6.67 & 33.58 & 1.75  \\
        kgS2S & 1.50  & 8.74 & 39.63 & 1.72  \\
        PGN & 1.84 & 9.80 & 39.89 & 1.98 \\
        kgPGN &  1.59  & 9.84 & 41.52 & 2.13 \\
        DeepCopy & 1.82 & 9.58 & 41.25 & 2.13 \\ \hline
        Ours, $\beta=0$  & \textbf{1.97}  & 10.86 & 44.21 & 2.41 \\
        Ours, $\beta=1$ & 1.02 & \textbf{15.40} & \textbf{58.88} & 2.48 \\
        Ours & 1.69 & 15.18 & 53.63 & \textbf{2.59}\\        
        w/o $\mathcal{L}_{sw}$ & 1.87 & 11.55 & 45.70 & 2.55 \\
        w/o $\mathcal{L}_{cp}$ & 1.58 & 9.66 & 40.84 & 1.75 \\      
        w/o $\mathcal{L}_{sw}$ \& $\mathcal{L}_{cp}$ & 1.76 & 10.71 & 42.62 & 2.25 \\   \hline      
    \end{tabular}
    \caption{Automatic evaluation results. The best results are highlighted in \textbf{bold}, and higher value denotes the better performance. BLEU and Dist scores are shown in  percentage. The results with * are taken from the DSTC7 paper. $\dag$ denotes the kgPGN model.}
    \label{tab:automatic_result}
\end{table}
Table \ref{tab:automatic_result} shows the automatic evaluation results. It can be seen that our model achieves better performance on all the metrics than all the baselines and the submissions to the DSTC7. This indicates that our proposed model can produce diversified responses that are smoothly connected to the source inputs. Moreover, ablation tests show that both the switch loss $\mathcal{L}_{sw}$ and the copy loss $\mathcal{L}_{cp}$ significantly boost the diversity scores, and the $\mathcal{L}_{cp}$ also helps to improve the scores of BLEU and PMI. This shows that learning to select which the decoding type should be used explicitly and encouraging our model to use the copy distributions preferentially are vital to improve the performance of our model. Moreover, the combination of two loss functions is necessary if the best performance is to be obtained. The results of the models by setting the probability in Eq. \ref{eq:switch} to 0 (convergent decoding) and 1 (divergent decoding) show that our models with $\beta=0,1$ achieve the highest BLEU and Dist-1,2, respectively. This indicates the $\beta$ can be effectively used to control the behavior of the generated responses.

\begin{figure}[htb]
  \begin{center}
     \includegraphics[width=7cm, height=5cm]{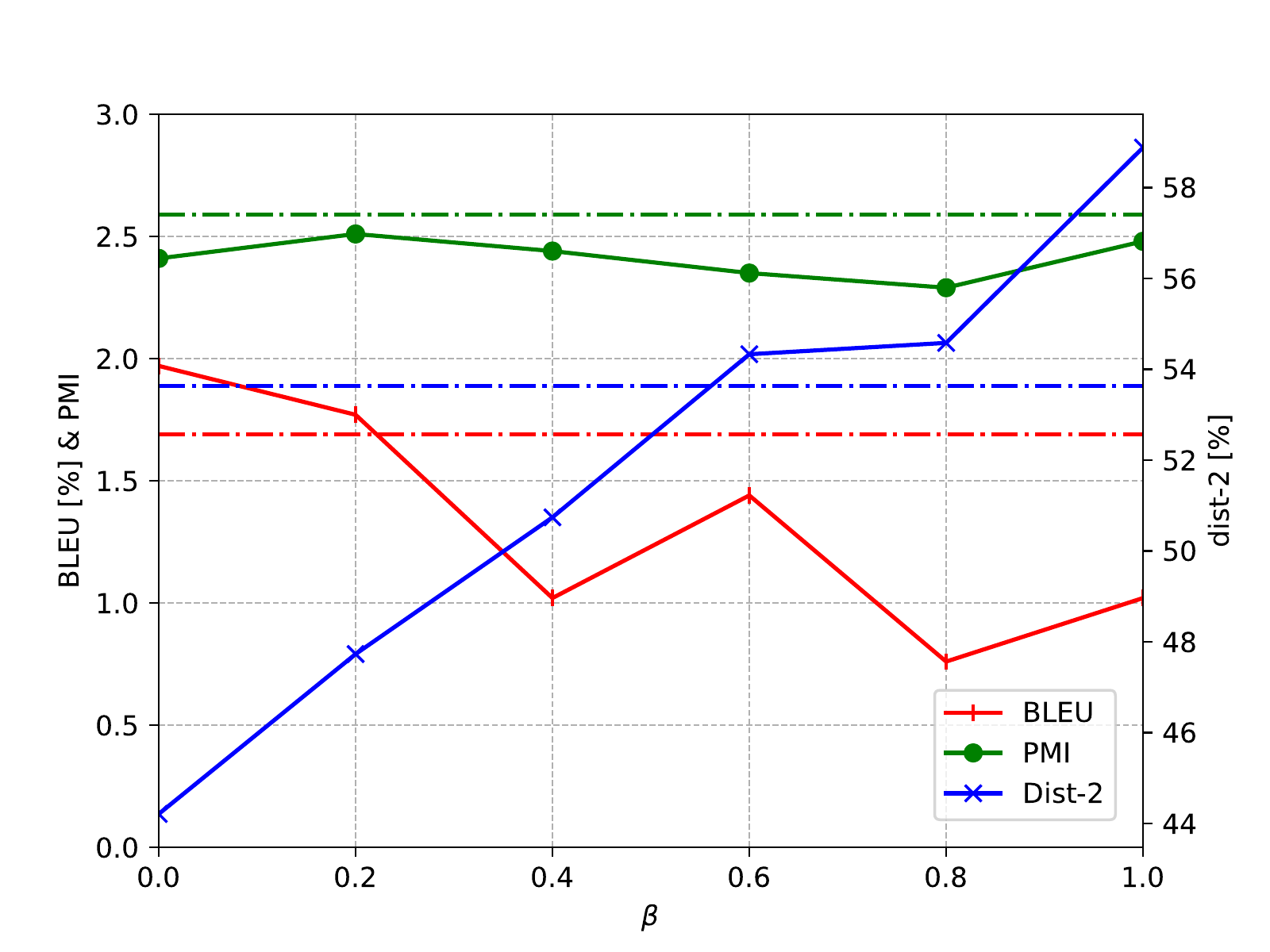}
    \caption{Effect for adjusting the $\beta$. Dash lines denote the results of proposed model that utilizes $\beta$ predicted by the decoding switcher.}
    \label{fig:eval_beta}
  \end{center}
\end{figure}

Furthermore, the effect of the probability $\beta$ on the generated responses was further evaluated. Specifically, we calculated the score of BLEU, Dist-2, and PMI corresponding to different $\beta$ values (see Figure \ref{fig:eval_beta}). According to the results, we can observe that BLEU decreases and Dist-2 increases when the $\beta$ increases. In other words, there is a trade-off relationship between relevance and diversity. This observation is in line with the previous work \cite{DBLP:journals/corr/abs-1811-08100}. Most importantly, any models do not achieve higher the score of PMI than our model using the dynamic value. This indicates that our decoding switcher is more effective to produce a response that can be connected to both context and facts smoothly.

\subsection{Human Evaluation}\label{sec:human_evaluation}
\begin{table}[htb]
    \centering
    \fontsize{9.3pt}{0.4cm}\selectfont
    \begin{tabular}{c|ccc} \hline   
        Model &  Appropriateness & Informativeness & Fluency \\ \hline
        S2S & 0.873 & 0.603 & 1.207 \\ 
        kgS2S &  0.950 & 0.620 & 1.257 \\
        PGN & 1.017 & 0.713 & 1.300 \\
        kgPGN & 1.087 & 0.890 & 1.237 \\ 
        DeepCopy & 0.960 & 0.900 & 1.230 \\ \hline
        Ours, $\beta=0$ & 1.040 & 0.783 & 1.153 \\
        Ours, $\beta=1$ & 1.020 & 0.876 & 1.100 \\        
        Ours & \textbf{1.243} & \textbf{1.100} & \textbf{1.310} \\  \hline
    \end{tabular}
    \caption{Human evaluation results.}
    \label{tab:human_evaluation}
\end{table}

\begin{table*}[htbt]
    \centering
    \fontsize{8pt}{0.4cm}\selectfont
    \begin{tabular}{c|p{7cm}|p{7cm}} \hline
         & \multirow{3}{7cm}{Before she was famous, Nicki Minaj was fired from approximately fifteen jobs. $\rightarrow$  So, she got hired 15 times. That' a positive. [...] $\rightarrow$ Helps being good looking.} & \multirow{3}{7cm}{Kyoto, the former capital of Japan, means ``capital city" and Tokyo means ``eastern capital" $\rightarrow$ New York means its a new city of York, and the word ``York" is celtic for the yew tree [...]}
         \\ Context & \\ & \\ \hline 
        S2S & Well I don’t care what she thinks she looks like & It's actually the only place. \\
        kgS2S & She looks a little like a girl. & New York is a new city ! \\
        GTTP & I think I like that. & No, that's just a myth. \\ 
        kgPGN & It's the best way to get a job. & I guess they have to make a new city of Japan. \\
        DeepCopy & No, Minaj gets a job. & And the Kyoto and New York .\\ \hline 
        Ours, $\beta$ = 0 & I didn't know that Minaj was fired 15 times. & Kyoto was the capital of Japan. \\ 
        Ours, $\beta$ = 1 & I love \textbf{Kanye} \textbf{Pitbull} \textbf{Kanye}. & I want to make a new city in \textbf{China}. \\
        Ours & I love \textbf{Kanye} and \textbf{Pitbull} ... and her sister is pretty hot. & New York is the most densely populated major city in America. \\  \hline 
    \end{tabular}
    \caption{Generated examples. The context is manually shortened to fit in the table, without significantly affecting meaning. The new topics that are drifted from the candidates of topics are shown in \textbf{bold}. Minaj, Kanye, and Pitbull are famous for an American rapper. The decoding switcher predicts the $\beta=0.85,0.12$ in the left and right examples, respectively.}
    \label{tab:response_example}
\end{table*}
As shown in Table \ref{tab:human_evaluation}, on the human evaluation, our model shows the highest performance on the all the metrics among all the models. Specifically, Appropriateness and Informativeness are significantly improved while retaining Fluency. This result indicates that our model can generate more meaningful responses that is related/appropriate to the conversation. Further observation shows that our model using the decoding switcher significantly outperforms our models with $\beta=0,1$ in all the metrics. This indicates that selecting the decoding type forcefully hurts response fluency and usefulness for the user. This verify the effects of our model using the decoding switcher. This is also in line with the score of PMI in the automatic evaluation in Table \ref{tab:automatic_result} and Figure \ref{fig:eval_beta}.

\section{Case Study}
Table \ref{tab:response_example} shows the two examples of the responses produced by all the models. From the results, our proposed model with the decoding switcher can generate more informative and contentful responses that are appropriate to the context than the other systems, while the baselines tend to talk about the current or previous topics passively and sometimes result in the redundant and unnatural conversation. These cases also show that the $\beta$ can be effectively used to flexibly control whether to converse about the exiting topics or drift the topic. 

We can observe that there are also some cases where our models with $\beta=0,1$ do not perform well. These our models sometimes generate the non-fluent response ($\beta=1$ in the left example) and the  non-appropriate response to the context ($\beta=0$ in the right example). By contrast, our model using the decoding switcher can generate more conversationally appropriate and fluent responses. 

\section{Related Work}
Sequence to sequence (seq2seq) approach has gained great attention in the field of dialogue generation. However, the problem of tending to generate generic and non-informative responses such as ``\textit{I don't know}" and ``Thanks" still remains \cite{DBLP:conf/coling/VougiouklisHS16,DBLP:conf/icann/YeCLHL18,DBLP:conf/acl/HuangKGx18}.

The incorporation of facts (unstructured knowledge) has been shown to be an effective way to solve these problems. \cite{DBLP:conf/aaai/GhazvininejadBC18} extended encoder-decoder model where the decoder is provided with an encoding of the context along with the facts encoding. To capture the multi-turn dialogue context and facts effectively, \cite{DBLP:journals/corr/abs-1902-01529} proposed a fact-based hierarchical seq2seq model. \cite{DBLP:conf/ijcai/LianXWPW19} proposed a knowledge selection mechanism where both prior and posterior distributions over knowledge were used. To explicitly output  informative words, \cite{DSTC7-SOTA} and \cite{DBLP:journals/corr/abs-1908-10731} proposed a model that enable the decoder to copy and attend tokens from any of the facts available as external knowledge in addition to the dialogue context. These works mainly focus on integrating facts based on a topic of input utterances into dialogue systems, which implicitly assume the topic to be kept on a dialogue and usually converse passively. However, this does not meet the common practice observed in our daily communication, and these systems have a difficulty to generate diverse responses that provide meaningful information proactively. To address this issue, we first incorporate the concept of the convergent and divergent thinking over both context and facts that can be observed in human thought method \cite{guilford1967nature} into fact-based dialogue system.

One of the most related works to ours may be \cite{DBLP:conf/acl/LiuFCRYL18}, who proposed a knowledge diffusion model (NKD) that can not only match the knowledge base (structured knowledge) for the input utterance but diffuse them to similar entities with another type of knowledge base, but we have following differences. Firstly, the NKD is designed to depend on the movie specific-domain knowledge base, making them not applicable for open-domain dialogue generation. Moreover, such knowledge requires a lot of works to build up. Meanwhile, we applied a very large scale and open-domain facts (unstructured knowledge that can be relatively easily constructed) to show the effectiveness of our model in real-world conversation. Secondly, the NKD cannot explicitly control whether to exhibit similar entities or not, while our model can flexibly control the behavior of responses by the convergent and divergent decoding. 

\section{Conclusion and Future Work}
This paper proposed a fact-based dialogue model with a convergent and divergent decoding that introduces a concept of a convergent and divergent thinking over both context and facts. In contrast to existing approaches, our model can flexibly control either conversing about the current topic or drifting the topic, which brings more diverse and proactive conversation. Evaluation results show that our method significantly outperforms state-of-the-art baselines in producing more appropriate, informative, and diverse responses. This paper shed some lights in exploring the potential of incorporation of the convergent and divergent thinking into dialogue systems. As for future work, we plan to apply a reinforcement learning to further improve the performance of the model. 

\bibliographystyle{named}
\bibliography{ijcai20}

\end{document}